\documentclass[conference]{IEEEtran}
\IEEEoverridecommandlockouts
\usepackage{cite}
\usepackage{amsmath,amssymb,amsfonts}
\usepackage{algorithmic}
\usepackage{graphicx}
\usepackage{textcomp}
\usepackage{xcolor}
\usepackage{multirow}
\usepackage{subcaption}
\usepackage{gensymb}
\def\BibTeX{{\rm B\kern-.05em{\sc i\kern-.025em b}\kern-.08em
    T\kern-.1667em\lower.7ex\hbox{E}\kern-.125emX}}
\begin{document}

\title{High Torque Density PCB Axial Flux Permanent Magnet Motor for Micro Robots}

\author{
\IEEEauthorblockN{
Jianren Wang\IEEEauthorrefmark{1}*,
Quanting Xie\IEEEauthorrefmark{1}*,
Jie Han\IEEEauthorrefmark{2},
Yang Zhang\IEEEauthorrefmark{2},\\
Christopher G. Atkeson\IEEEauthorrefmark{1},
Abhinav Gupta\IEEEauthorrefmark{1},
Deepak Pathak\IEEEauthorrefmark{1},
Yonatan Bisk\IEEEauthorrefmark{1}
}
\IEEEauthorblockA{\IEEEauthorrefmark{1}Robotics Institute, Carnegie Mellon University, PA, USA \\
\{jianrenw, quantinx, cga, abhinavg, dpathak, ybisk\}@andrew.cmu.edu}
\IEEEauthorblockA{\IEEEauthorrefmark{2}R\&D, Microbot Motor, Inc., Suzhou, China \\
\{jack.li, thomas.zhang\}@microbot-motor.cn}
\thanks{* Equal contribution, order decided by coin flip.}
}

\maketitle

\begin{abstract}
Quasi‑direct‑drive (QDD) actuation is transforming legged and manipulator robots by eliminating high‑ratio gearboxes, yet it demands motors that deliver very high torque at low speed within a thin, disc‑shaped joint envelope. Axial‑flux permanent‑magnet (AFPM) machines meet these geometric and torque requirements, but scaling them below a 20mm outer diameter is hampered by poor copper fill in conventional wound stators, inflating resistance and throttling continuous torque. This paper introduces a micro‑scale AFPM motor that overcomes these limitations through printed‑circuit‑board (PCB) windings fabricated with advanced IC‑substrate high‑density interconnect (HDI) technology. The resulting 48‑layer stator—formed by stacking four 12‑layer HDI modules—achieves a record $45\%$ copper fill in a package only 5mm thick and 19mm in diameter. We perform comprehensive electromagnetic and thermal analyses to inform the motor design, then fabricate a prototype whose performance characteristics are experimentally verified.
\end{abstract}


\section{Introduction}

In recent years, quasi‑direct‑drive (QDD) actuation has spurred rapid progress across legged machines and manipulators, including quadrupeds~\cite{bledt2018cheetah}, bipeds~\cite{saloutos2023design}, lightweight robot arms~\cite{zhao2023light}, and high‑performance grippers~\cite{bhatia2019direct}. By placing a low‑ratio transmission—or no gearbox at all—between a torque‑dense motor and the joint, QDD robots gain high back‑drivability, low reflected inertia, minimal friction and backlash, and therefore superior force control bandwidth and energy efficiency. These advantages, however, come with two key design constraints: 1) High torque at low speed is required because the gear ratio is small. 2) The joint envelope is usually disc‑shaped—large radius but shallow height—demanding a flat, high‑torque motor topology.

Axial‑flux permanent‑magnet (AFPM) machines naturally satisfy both constraints. Their torque scales with the cube of the rotor radius (versus the square for radial‑flux machines), enabling high torque at low rpm, while their thin pancake geometry fits easily inside robot joints and simplifies heat extraction~\cite{seo2010design}.

However, miniaturizing AFPM for micro‑robotic joints ($\leq20mm$ outer diameter) is difficult, chiefly because conventional wound stators cannot accommodate the required slot fill and insulation thickness when the wire diameter drops below $\sim 50\mu m$. Low copper fill ($<35\%$) leads to excessive resistance, poor continuous torque, and thermal bottlenecks.

To overcome these manufacturing limitations, printed circuit board (PCB)-based winding technology has emerged as an attractive alternative due to its scalable production, cost-effectiveness, and compatibility with microfabrication techniques, making them ideal for micro-robotic applications~\cite{wu2012design}. Yet off‑the‑shelf PCB motors still suffer from low copper fill factors and therefore limited torque density~\cite{10475522}. Addressing this shortcoming is critical, as it directly influences the motor's achievable torque and efficiency, both vital metrics in high-performance robotic actuators.

In this paper, we present a novel high torque-density PCB-based AFPM motor explicitly designed for micro-robotic applications. We leverage advanced IC substrate fabrication techniques combined with High-Density Interconnect (HDI) technology to significantly enhance \textbf{copper fill} factors, achieving values exceeding $45\%$—a marked improvement over traditional PCB-based designs. Our design involves stacking multiple 12-layer PCB modules interconnected via sophisticated HDI methods to form a 48-layer stator only $5mm$, resulting in substantial improvements in both electrical and thermal performance.

The motor design (Section~\ref{sec:design}) is guided by electromagnetic simulations and thermal performance evaluations (Section~\ref{sec:sim}), followed by prototyping and validation through real-world experiments (Section~\ref{sec:exp}). Our prototype motor surpasses commercially available alternatives—providing \textbf{higher torque} under \textbf{low-speed} operation while occupying substantially less volume. These attributes position our PCB AFPM motor as a promising actuator solution for the next generation of micro-robotic systems, enabling more efficient, compact, and reliable robotic platforms.

\section{Design}
\label{sec:design}

\begin{figure*}[ht]
    \centering 
    \includegraphics[width=0.9\linewidth]{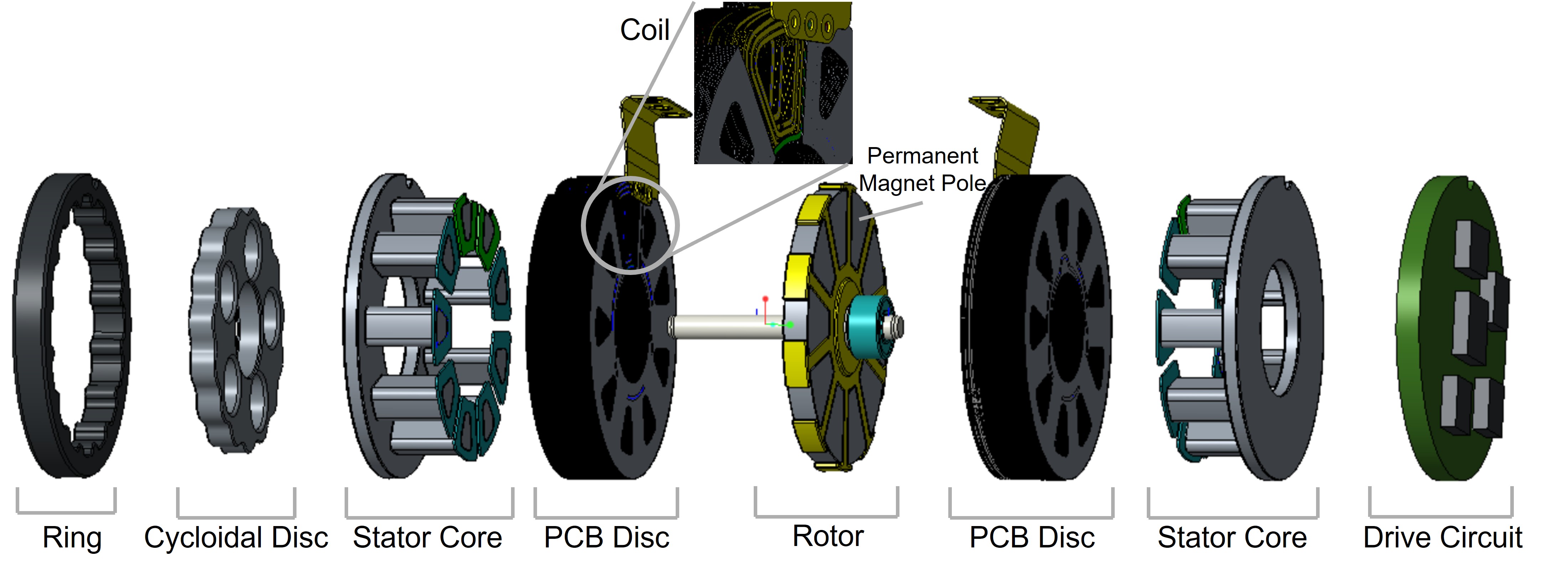} 
        \caption{Exploded view of double stator-single rotor PCB Axial Flux Permanent Magnet Motor}
    \label{fig:exploded}
\end{figure*}

Fig~\ref{fig:exploded} shows an exploded view of the proposed two-stator, one-rotor axial-flux internal rotor (\textit{i.e.} AFIR-AFPM motor). Unlike most PCB motors—which eliminate the steel core—we retain the core alongside the PCB windings. Since most robotic applications operate at low speeds, the iron losses are minimal and can be neglected. Moreover, including the steel core significantly enhances the magnetic flux in the motor.

\begin{table}[h]
    \centering
    \caption{Main Design Parameters OF The Proposed Motor}
    \begin{tabular}{|c|c|}
    \hline
        Pole Pairs & 5 \\ \hline
        Outer diameter (mm) & 19 \\ \hline
        Inner diameter (mm) & 7.0 \\ \hline
        Axial length of air gap (mm) & 0.25\\ \hline
        Axial length of PCB stator (mm) & 6.0\\ \hline
        Number of virtual slots & 9 \\ \hline
        Thickness per PCB layer (mm) & 0.06\\ \hline
        Number of single PCB layers & 48\\ \hline
        Turns of coil / Number of PCBs connected in series & 3\\ \hline
        Number of parallel branches & 2\\ \hline
        Trace width (mm) & 0.36 \\ \hline
        Copper fill factor & $\ge 45\%$\\ \hline
        Nominal  voltage (V) & $24\sim48$ \\ \hline
    \end{tabular}
    \label{tab:design}
\end{table}

\begin{figure}
    \centering
    \includegraphics[width=\linewidth]{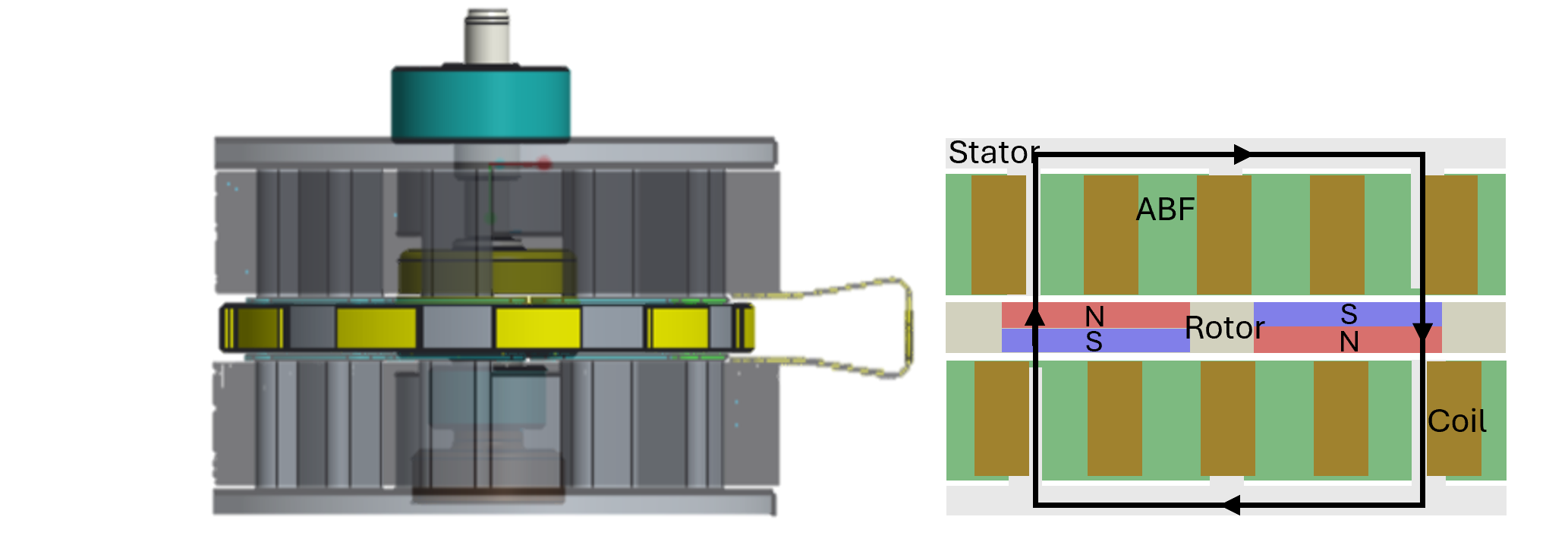}
    \caption{Flux path of the PCB-based AFPM Motor}
    \label{fig:flux_direction}
\end{figure}

Figure~\ref{fig:flux_direction} presents a side view of the proposed PCB-based axial-flux permanent-magnet (AFPM) motor, along with a schematic illustration of its magnetic circuit. In the diagram, the arrows indicate the intended closed magnetic flux loop: the flux originates from the north poles of the surface-mounted rotor magnets, traverses the air gap into the planar PCB windings, spreads radially through the axial back-iron (ABF) return plates, and finally re-enters the adjacent south poles. This configuration differs from conventional AFPM designs, where a piece of iron typically connects the north and south poles, creating two separate flux paths. In contrast, the proposed design eliminates this intermediate iron bridge, resulting in a thinner rotor structure and reduced magnetic flux leakage.

\begin{figure}[h]
    \centering
    \begin{subfigure}[b]{0.23\textwidth}
    \includegraphics[width=\textwidth]{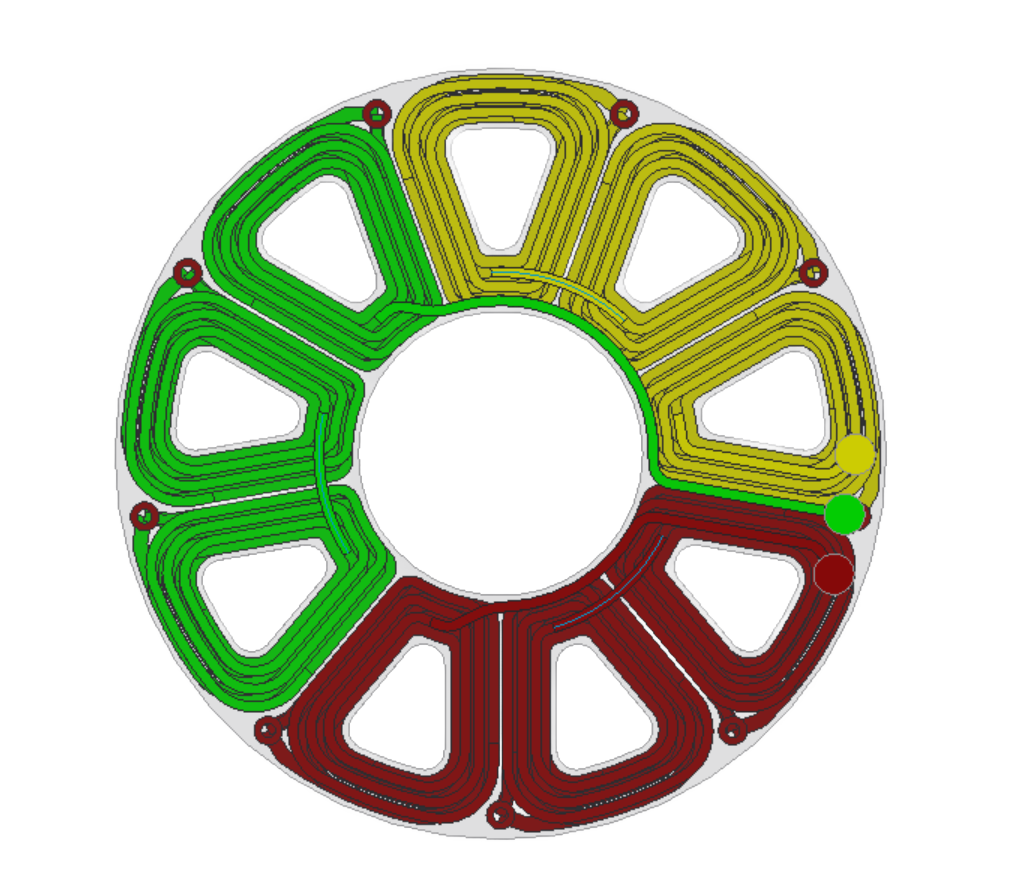}
    \caption{Chosen winding layout.}
    \label{fig:winding}
    \end{subfigure}
    \hfill
    \begin{subfigure}[b]{0.23\textwidth}
    \includegraphics[width=\textwidth]{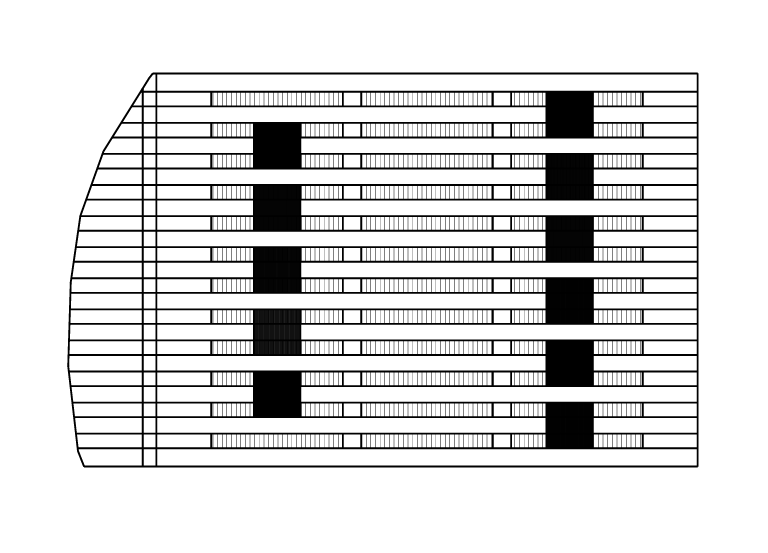}
    \caption{Details of the High Density Interconnect (HDI) layout.}
    \label{fig:hdi}
    \end{subfigure}
\end{figure}

In this paper, we propose, fabricate, and analyze a $14mm \times 19mm$ motor. Table~\ref{tab:design} summarizes the motor’s specifications. We evaluated several PCB layouts, but with a stator core in place, the performance differences were minimal. Therefore, for ease of fabrication, we adopted a standard PCB layout (see Fig~\ref{fig:winding}). Fig~\ref{fig:hdi} shows how a 12-layer module is interconnected using HDI technology—horizontal strips represent the copper layers, and black rectangles denote the interlayer via “stacks” that form the multi-layer coil paths. As indicated in Table~\ref{tab:design}, using IC substrates instead of conventional PCB materials significantly increases the copper fill factor. Notably, our design consists of four series-connected 12-layer modules (totaling 48 layers), which is substantially more than the 6 layers typically found in other PCB motors~\cite{wang2021design}.

\section{Electromagnetic Analysis and Thermal Analysis}
\label{sec:sim}

The motor presented in this work is specifically engineered for robotic applications that demand high torque at low rotational speeds and frequent operation under stall conditions. The design methodology follows a structured approach grounded in both electromagnetic~\cite{zhu2000finite, zhou2002field} and thermal analysis~\cite{roy2024heat}.

\begin{figure}
    \centering
    \includegraphics[width=0.9\linewidth]{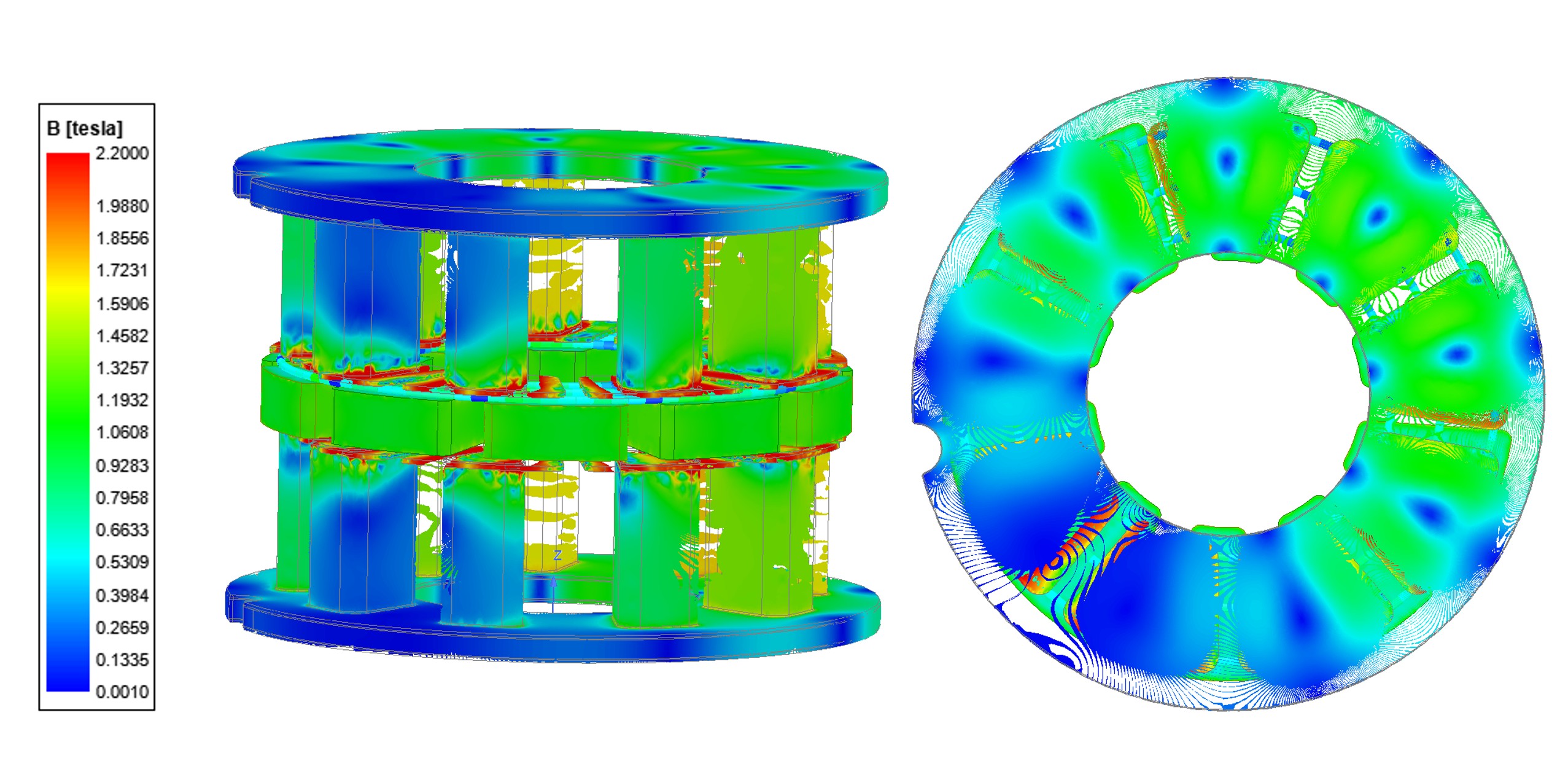}
    \caption{No-load magnetic flux density (B)}
    \label{fig:magnetic}
\end{figure}

Similar to~\cite{quan2025design}, the design process begins with an evaluation of magnetic flux density to guide the optimization of the yoke geometry, magnetic material selection, magnet configuration, and coil winding parameters. To prevent magnetic saturation in the stator teeth and yoke, the flux density in the core is constrained to remain below the material's saturation limit. This is expressed as: $B_{\text{core}} = \frac{\Phi}{A_{\text{core}}} \leq B_{\text{sat}}$ where \(\Phi\) is the magnetic flux, \(A_{\text{core}}\) is the cross-sectional area of the core (teeth or yoke), and \(B_{\text{sat}}\) is the saturation flux density of the core material. Figure~\ref{fig:magnetic} presents the estimated no-load magnetic flux density from finite element analysis (FEA) performed using Ansys Maxwell~\cite{ANSYS_Maxwell_2025R1}. A high-saturation magnetic material (2.4 T) is selected to enhance magnetic flux capacity and prevent saturation under operational loads.

\begin{figure}
    \centering
    \includegraphics[width=0.9\linewidth]{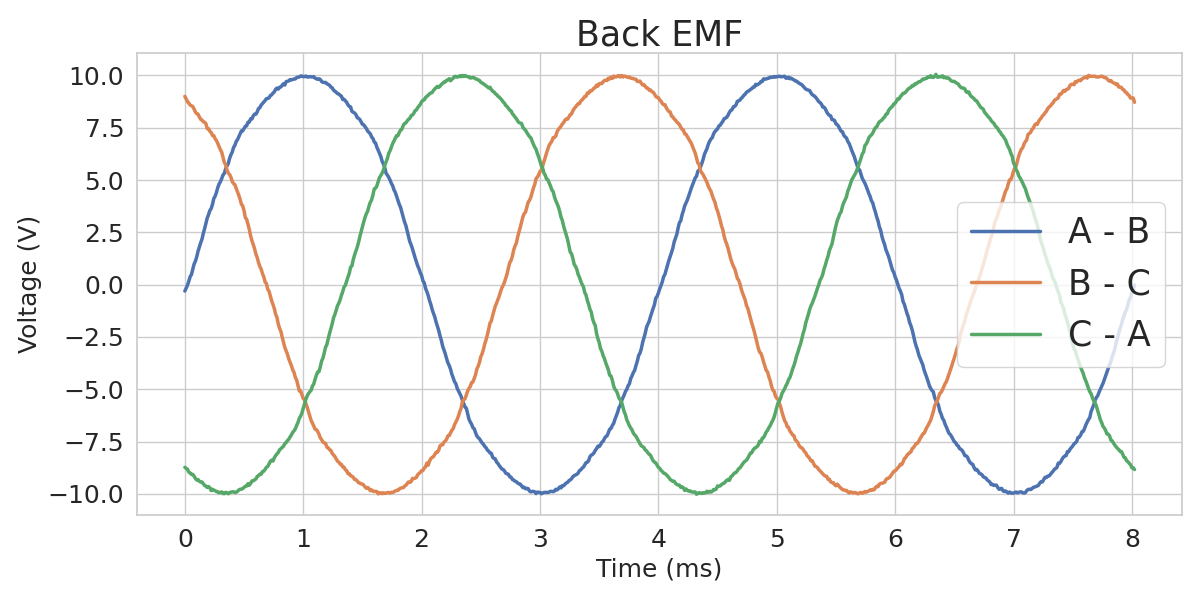}
    \caption{Simulated back EMF at 3000 rpm.}
    \label{fig:back_emf_sim}
\end{figure}

Subsequently, the back electromotive force (EMF) is simulated under both no-load and load conditions to verify that the combined back EMF and resistive voltage drop (IR drop) remain within the nominal input voltage limit. Figure~\ref{fig:back_emf_sim} shows the back EMF waveform at 3000 rpm, with a peak amplitude of 9.97 V. The waveform demonstrates a well-formed sinusoidal shape, indicating good electromagnetic symmetry and balanced winding distribution.

\begin{figure}
\centering
\includegraphics[width=0.9\linewidth]{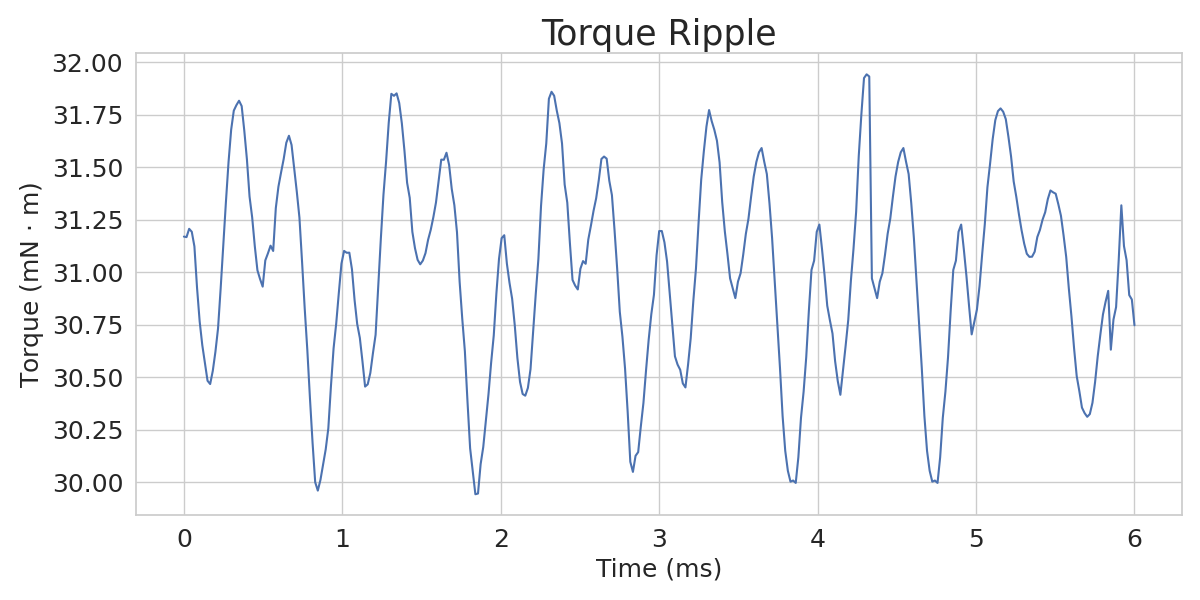}
\caption{Torque ripple at 2000 rpm, 30.7 mNm output torque.}
\label{fig:torque_rippling}
\end{figure}

\begin{figure}
\centering
\includegraphics[width=0.9\linewidth]{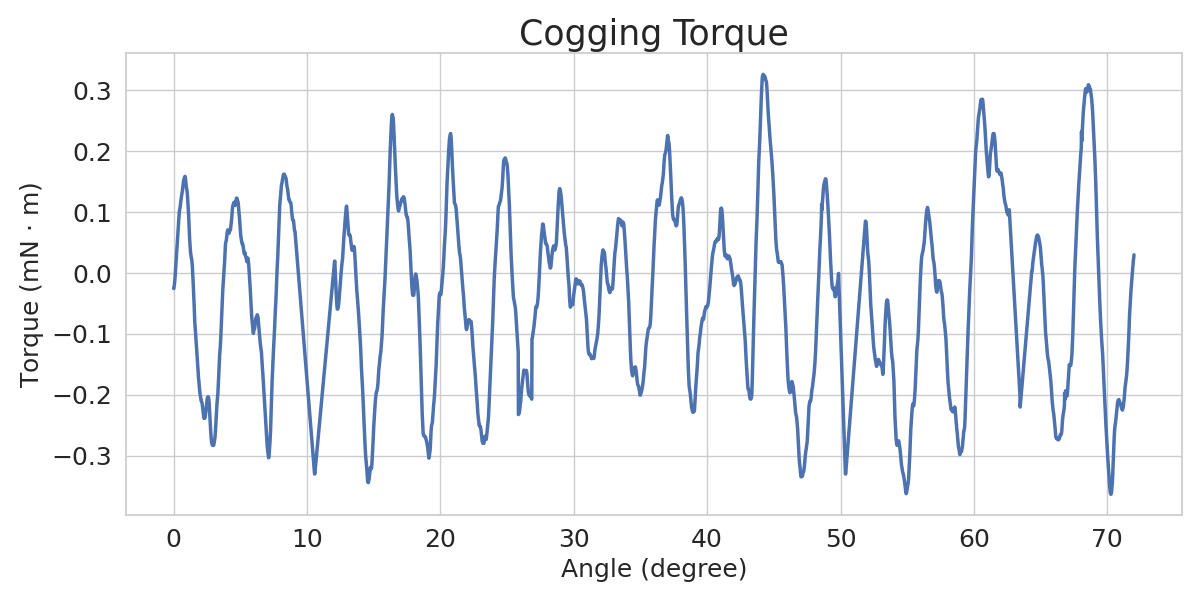}
\caption{Simulated cogging torque.}
\label{fig:cogging_torque}
\end{figure}

To minimize torque ripple during steady-state operation, the magnet arc and stator slot geometry are iteratively tuned based on torque waveform analysis, ensuring a smooth air-gap flux distribution. As shown in Figure~\ref{fig:torque_rippling}, the motor achieves a torque output of 30.7 mNm at 2000 rpm, with a ripple magnitude controlled below $6\%$, indicating stable torque performance. Cogging torque is also evaluated and reduced through further structural refinement, such as optimized pole-slot alignment, slot skewing, or magnet edge shaping. The simulated cogging torque, depicted in Figure~\ref{fig:cogging_torque}, confirms a low level of cogging torque; however, it is acknowledged that manufacturing tolerances may cause higher cogging levels in practice.

\begin{figure}
\centering
\includegraphics[width=0.9\linewidth]{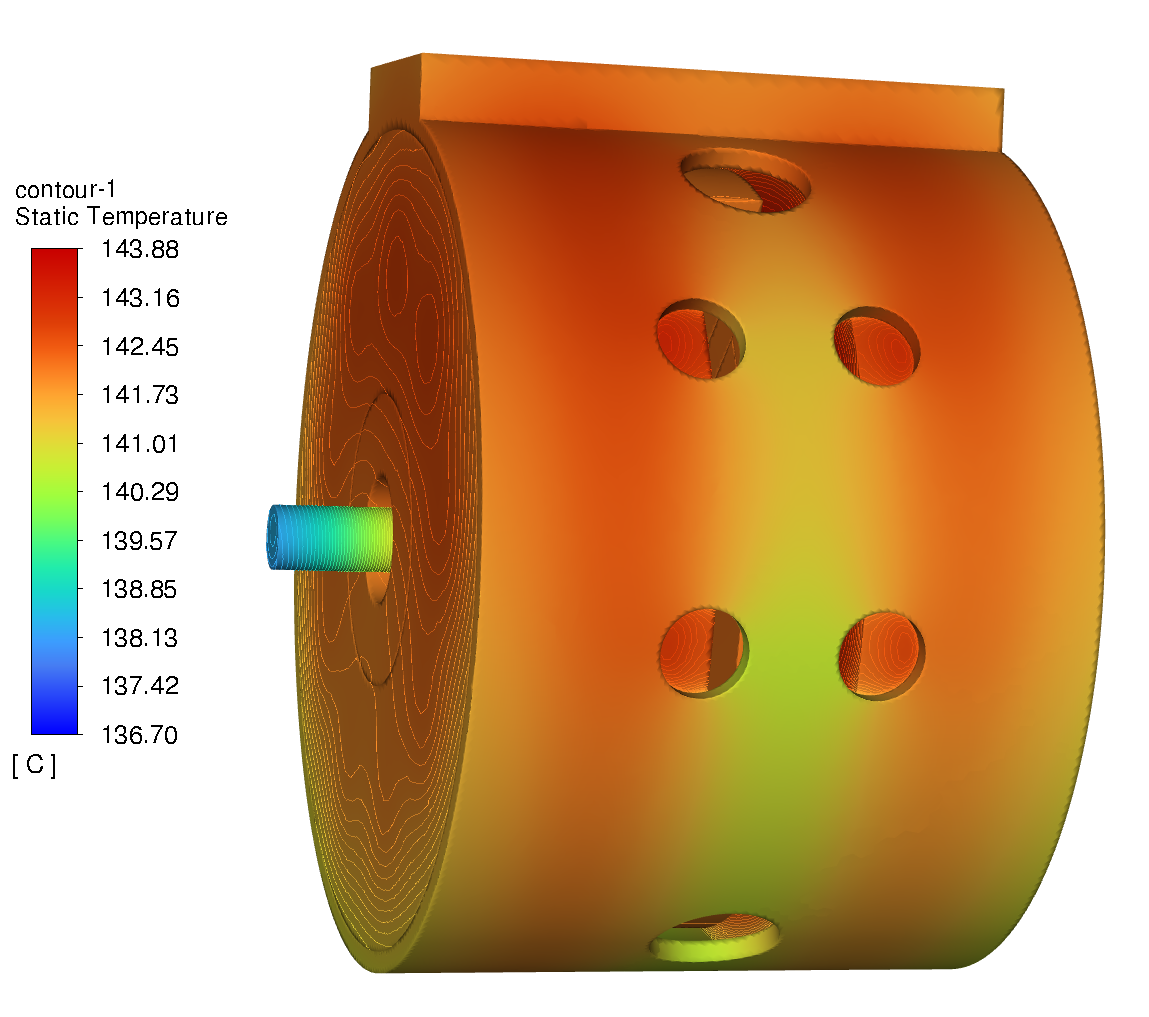}
\caption{Temperature distribution under stall conditions.}
\label{fig:thermal}
\end{figure}

Finally, thermal performance is evaluated using Ansys Fluent~\cite{ANSYS_Fluent_2025R1} to determine the allowable continuous stall torque. The thermal behavior is governed by the heat conduction equation:$\rho c_p \frac{\partial T}{\partial t} = \nabla \cdot (k \nabla T) + Q$, where: \(\rho\) is the material density, \(c_p\) is the specific heat capacity, \(T\) is the temperature, \(k\) is the thermal conductivity, and \(Q\) represents the volumetric heat generation (mainly due to copper losses). Under steady-state conditions (\textit{i.e.}, when \(\frac{\partial T}{\partial t} = 0\)), our simulation (see Figure~\ref{fig:thermal}) predicted a peak motor temperature of \(143{\degree C}\) under 8W thermal load. This temperature remains within the safety margins of the selected materials: the insulation system is rated for operation up to $200^{\circ}\mathrm{C}$, and the permanent magnets are specified for thermal stability up to $180^{\circ}\mathrm{C}$. These specifications ensure reliable thermal performance under continuous stall conditions.

% ci tong mi du queue ren zui jia de ci tie xing zhuang (houdu / jiaodu), dao ci cai liao (chi bu e bu ing zhuang), que ding zui jia she ji

% back emf que ding kong zai zui gao zhuan su / fu zai gong kuang xia fan dian shi jia shang ya jiang bu neng chao guo e ding dian ya

% torque rippling e ding gong kuang bu neng tai da: $6\%$, qi xi ci chang bu zheng xian le, niu ju da yi dian, ci gang jiao du, cao kou kuan du, ding zi ci chang ou he, hui bu yi yang

% cogging torque: jie jue di su xia de zhuan ju bo dong, zhuan zi ci chang, cao kou

% themeral: ju yuan cai liao de bao hu, ju yuan ceng nai wen yao qiu, 200 160/170; 180 150
% tong man lv yi ding, zha shu mei you yong, ci chang geng bao he yi dian, zha shu bu bian
% gao su xing neng jin yi bu cha tie sun ci la li zhuan ju bo dong

% 1) dian ya bu chao xian dia ya fu zhi 

% xian dia ya fu = sqrt(2) xian you xiao = sqrt (3) xiang you xiao
% xian dia ya fu + I(xian dia liu)R(xian dian zu = 2 xiang dian zu) < 48

% 2) zheng xian, fu li ye fen xi, xie bo zhan bi, qi xi ci mi ce bu dao

% We also optimize the slot opening, magnetic 

% bi jiao xiao, you hua le cao kou kuan du, you hua le ci ji jiao du.

%shi ce zhuan su hui hen di

%wen du yun xu fan wei nei zui hao zhuan ju

% 180 ci tie
% 200 du cai liao

\section{Experiment}
\label{sec:exp}

\begin{figure*}[h]
    \centering 
    \includegraphics[width=0.9\textwidth]{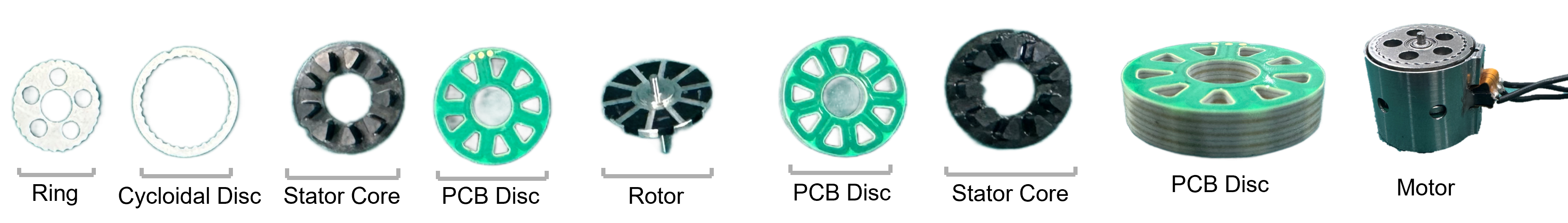} 
        \caption{Prototype Machine. Parts (Left), PCB Disc (Middle), Assembled Motor (Right)}
    \label{fig:prototype}
\end{figure*}

\begin{figure}
    \centering
    \includegraphics[width=0.9\linewidth]{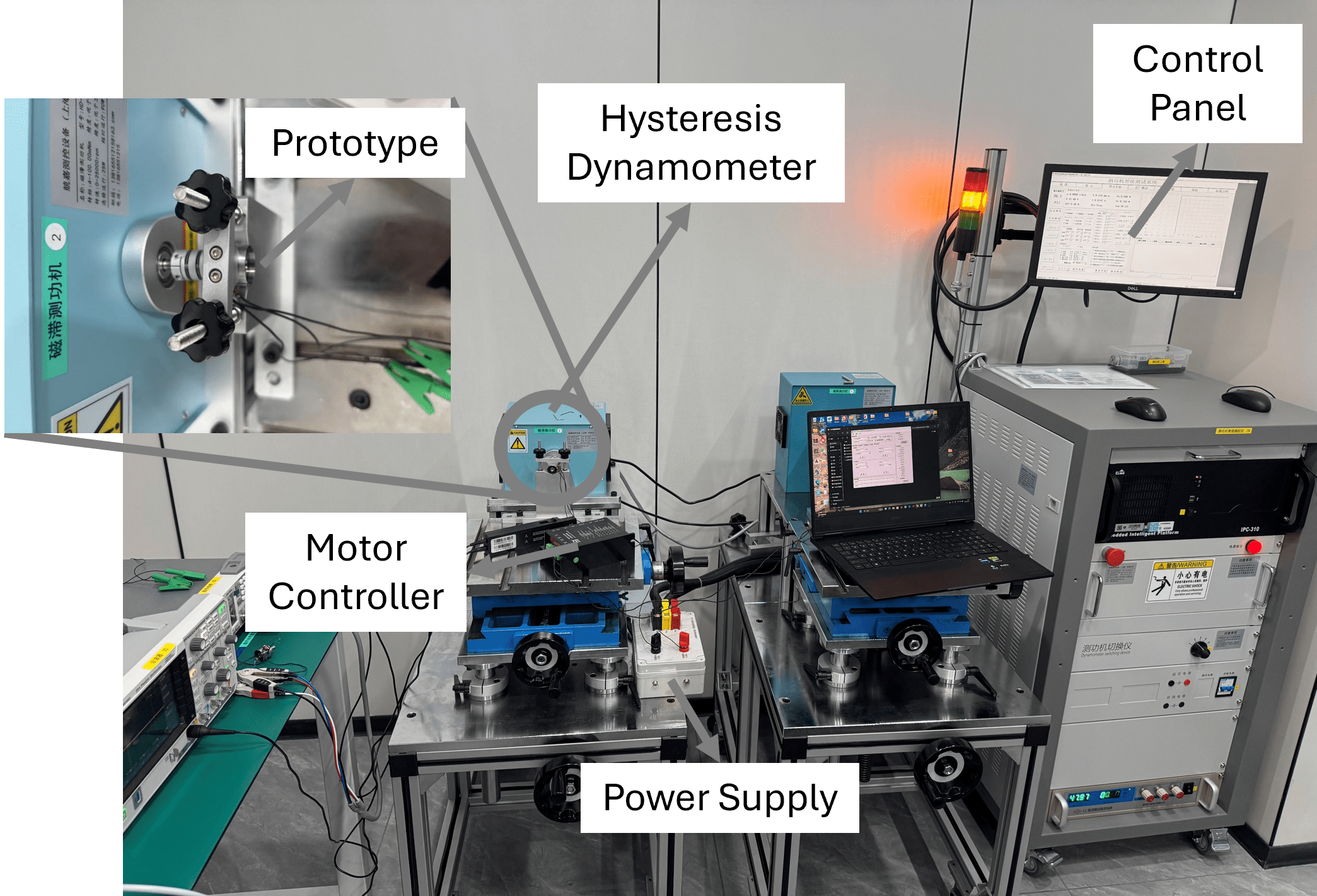}
    \caption{The experiment platform includes a hysteresis dynamometer, the prototype motor, a motor controller, a prime mover, and instruments for measuring voltage and current.}
    \label{fig:experiment}
\end{figure}

\begin{figure}
    \centering
    \includegraphics[width=0.9\linewidth]{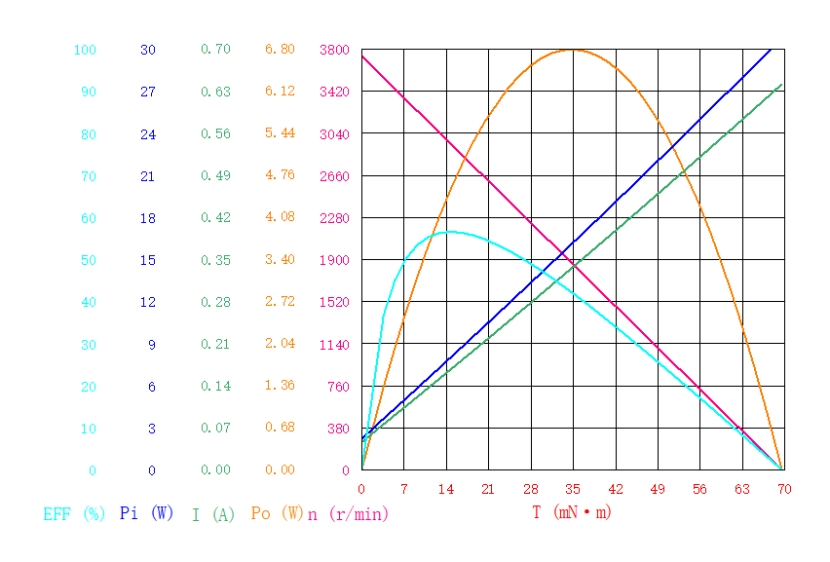}
    \caption{Output characteristics.}
    \label{fig:performance}
\end{figure}

\begin{figure}
    \centering
    \includegraphics[width=0.9\linewidth]{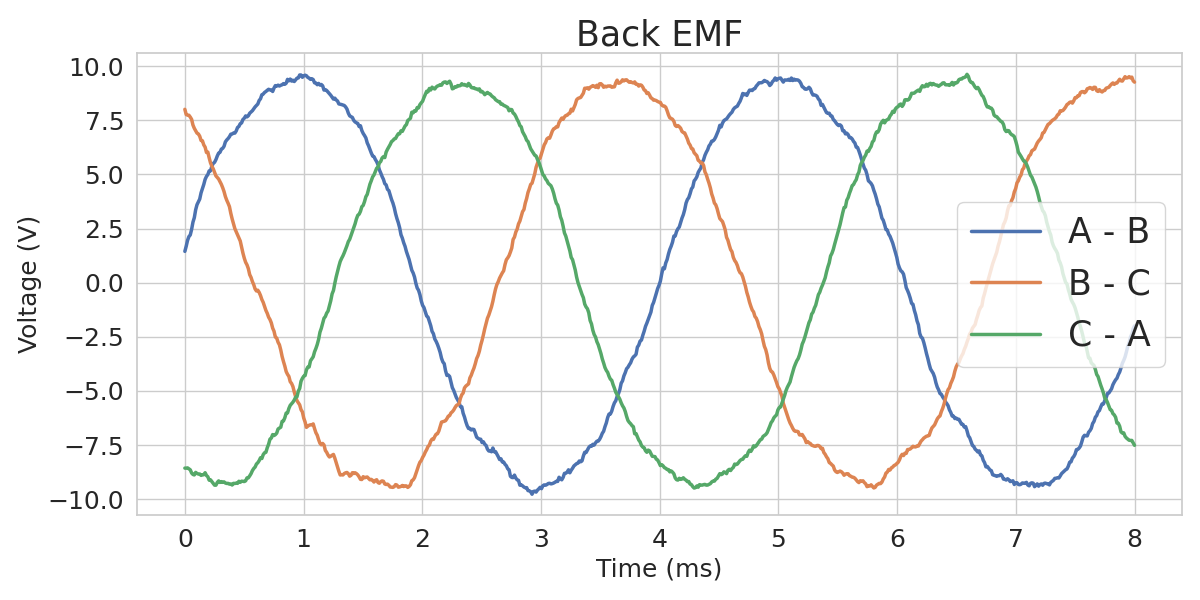}
    \caption{Actual back EMF under 3000rpm.}
    \label{fig:back_emf_real}
\end{figure}

\begin{figure}
    \centering
    \includegraphics[width=0.75\linewidth]{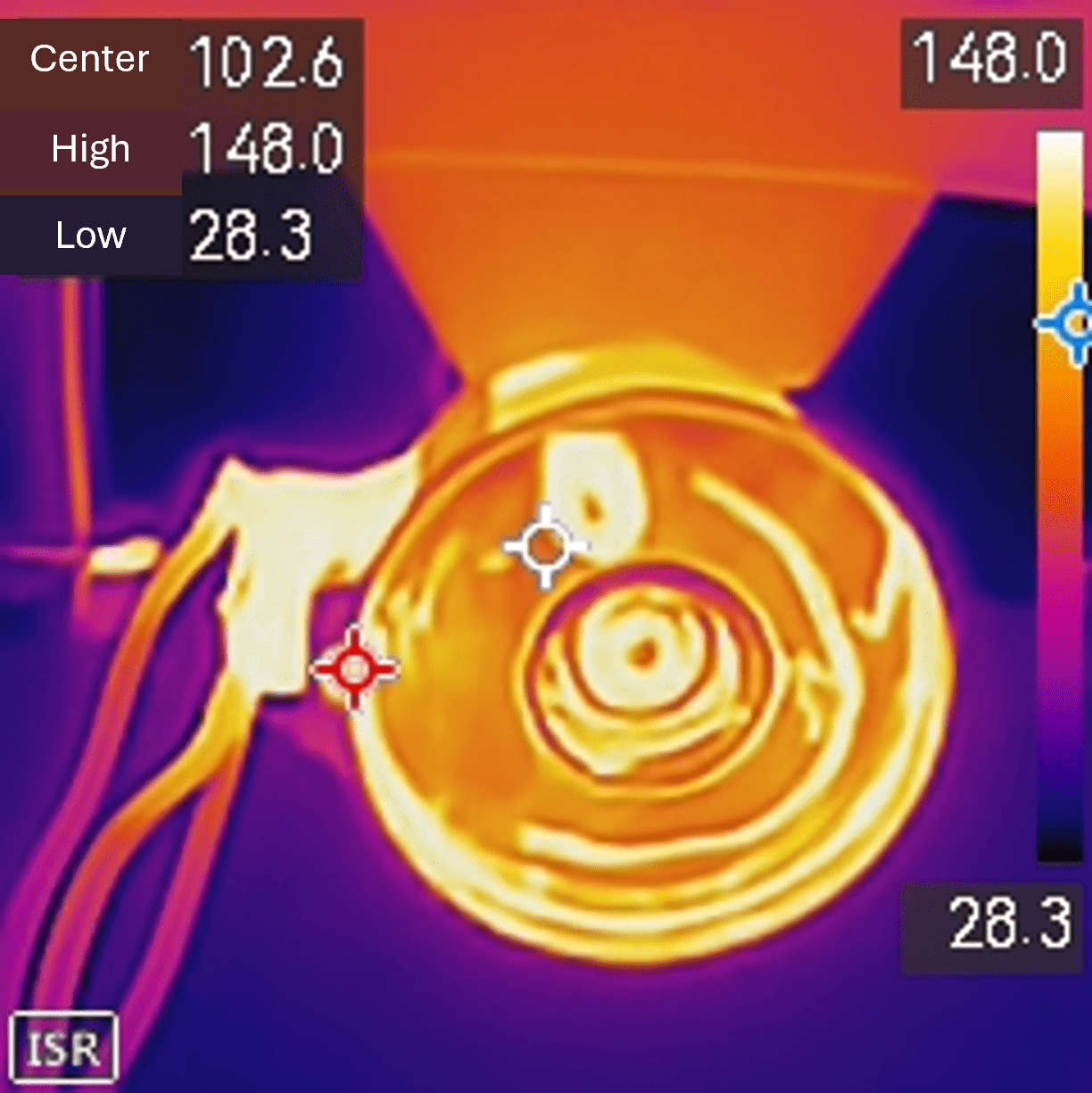}
    \caption{Thermal image under stall conditions.}
    \label{fig:thermal}
\end{figure}

A prototype machine was fabricated and tested (see Figure~\ref{fig:prototype}). The left panel illustrates the individual components, presented from left to right: outer ring (housing support), cycloidal disc, stator core (soft magnetic composite), PCB winding disc, rotor hub with embedded permanent magnets, second PCB winding disc, and opposing stator core. The middle panel shows the laminated stack of PCB windings. The right panel displays the fully assembled motor, enclosed in a steel housing with integrated wiring. Together, these panels highlight the compact, modular architecture realized in the hardware implementation.

We evaluate the performance of our prototype with the experimental setup (Figure~\ref{fig:experiment}). The experiment platform includes a hysteresis dynamometer, the prototype motor, a motor controller, a prime mover, and instruments for measuring voltage and current.

\begin{table}
  \centering
    \caption{PERFORMANCE OF PROTOTYPE MACHINE}
    \begin{tabular}{|c|c|c|}
        \hline
            Specs & Maxon & Ours \\  \hline 
            Mass w. gear 
            (gram) & 32 & 25\\ \hline
            No load speed (rpm) & 11100 & 5000 \\ \hline
            Nominal speed (rpm) & 8520 & 2000 \\ \hline
            Nominal torque (mNm) & 28 & 30.7 (Cold)/21.5 (Hot)\\ \hline
            Nominal current (A) & 0.885 & 0.96 \\ \hline
            Stall torque (mNm) & 126 &  158\\ \hline
            Continuous stall torque (mNm) & 20.0 & 23.4 \\ \hline
            Stall current (A) & 5.67 & 5.95 \\ \hline
            Continuous stall current (A) & - & 0.86 \\ \hline
            Terminal Resistance ($\Omega$) & 6.34 & 4.70\\ \hline
            Unbalance 3-phase resistance ($\%$) & - & $\leq 2 \%$ \\ \hline
            Unbalance 3-phase inductance ($\%$) & - & $\leq 1 \%$ \\ \hline
            Terminal inductance (mH) & 1.16 & 3 \\ \hline
            Torque constant (mNm/A) & 30.6 & 32.0 \\ \hline
            Speed constant (rpm/V) & 313 & 298 \\ \hline
            Rotor inertia (gcm$^2$) & 6.82 & 1.3 \\ \hline
            Cogging torque (mNm) & - &  $\leq 2$ \\ \hline
            Max. efficiency (\%) & 83 & 60 \\ \hline
            Stall torque density (Nm/$cm^3$) & 3.29 (Hot) & 5.32 (Hot)\\ \hline
    \end{tabular}
    \label{tab:experiments}
\end{table}

We compare our motor against a state-of-the-art commercially available miniature motor~\cite{maxon}, which serves as a relevant benchmark despite not being a one-to-one comparison. Our motor is specifically designed for robotic actuators that require high torque at low speeds, whereas the Maxon motor operates at relatively higher speeds. Nevertheless, it represents the closest available reference point for contextualizing the design intent and performance positioning of our prototype.

As shown in Table~\ref{tab:experiments}, our motor achieves a stall torque of 158 mNm and a continuous stall torque of 23.4 mNm, representing a $25.4\%$ and $16.4\%$ improvement, respectively, over the Maxon reference. Moreover, our design achieves this performance in a package approximately $25\%$ smaller in volume, resulting in a $61.7\%$ increase in stall torque density. This performance advantage is particularly meaningful for robotic actuation, where frequent operation near or at stall is common. In addition, our motor exhibits a higher torque constant ($K_t$) and a lower nominal speed compared to the Maxon motor, further underscoring its suitability for applications requiring high torque at low speeds—especially below 1500 rpm.

Figure~\ref{fig:performance} illustrates the motor's output characteristics under a $25\%$ duty cycle, corresponding to an effective input voltage of 12V (i.e., $48V \times 25\% = 12V$). The plot reports efficiency (\%), input power (W), output power (W), current (A), and rotational speed (rpm) as functions of output torque (mNm). The approximately linear relationships between speed–torque and current–torque facilitate straightforward integration into control systems. Notably, the motor achieves a substantial stall torque while maintaining high efficiency at moderate loads—an important advantage for robotic actuators, which frequently operate under stall or near-stall conditions.

We also measured the actual back EMF of the motor. As shown in Figure~\ref{fig:back_emf_real}, the experimental results closely match the simulation at 3000rpm, exhibiting a peak amplitude of approximately 9.48V—within 5\% of the simulated value. The waveform also demonstrates a well-formed sinusoidal shape, which is beneficial for modern motor control strategies such as field-oriented control (FOC).

Finally, the motor’s thermal performance was experimentally evaluated under stall conditions. A thermal image of the motor was captured using an infrared thermal camera, as shown in Figure~\ref{fig:thermal}. With a heat dissipation of 8W, the measured peak temperature reached $148^{\circ}\mathrm{C}$, closely matching the simulated result. Both the insulation system and magnetic materials remain within their rated thermal limits under these operating conditions, confirming the motor's suitability for continuous stall operation.

\section{Conclusion}

This work presents a high-torque-density, PCB-based axial flux permanent magnet (AFPM) motor tailored for micro-robotic applications, with particular emphasis on delivering high torque at low rotational speeds and supporting frequent stall operation. Leveraging integrated circuit (IC) substrates and high-density interconnect (HDI) technology, the 48-layer stator—constructed by stacking four 12-layer HDI modules—achieves a record copper fill factor of 45\% within a compact form factor measuring only 5mm in thickness and 19mm in diameter. The motor design is thoroughly optimized via simulation and validated through real-world experimental testing. We hope this work will inspire further development of PCB-based AFPM motors, particularly in the domain of micro-robotics.

\section*{Acknowledgment}

We would like to thank Wenbo Xu from Microbot Motor, Kenny Shaw, Andrew Wang, Tony Tao from CMU, Hejia Zhang, FNU Abhimanyu, Octavian Donca, Mohan Kumar Srirama, Tanmay Agarwal from Skild AI, Hang Zhao, Hanyang Zhou, Zeyi Yang from Galaxea AI for fruitful discussions.

\bibliographystyle{ieeetr}

\bibliography{reference.bib}
%}

\end{document}